%% file: main.tex
\documentclass{article}

\PassOptionsToPackage{numbers, compress}{natbib}
\usepackage[preprint]{neurips_2023}

\usepackage[utf8]{inputenc} % allow utf-8 input
\usepackage[T1]{fontenc}    % use 8-bit T1 fonts
\usepackage[hidelinks]{hyperref}       % hyperlinks
\usepackage{url}            % simple URL typesetting
\usepackage{booktabs}       % professional-quality tables
\usepackage{amsfonts}       % blackboard math symbols
\usepackage{nicefrac}       % compact symbols for 1/2, etc.
\usepackage{microtype}      % microtypography
\usepackage{xcolor}         % colors

\usepackage{bm}
\usepackage{amsmath}
\usepackage{amssymb}
\usepackage{multirow}
\usepackage{subfigure}
\usepackage{array}
\usepackage{amsthm}
\usepackage{rotating}
\usepackage{pifont}
\usepackage{wrapfig}

\newcommand{\tabref}[1]{Table \ref{#1}}
\newcommand{\secref}[1]{Section \ref{#1}}

\newcommand{\appendixref}[1]{Appendix}

\title{An Empirical Evaluation of \\ Temporal Graph Benchmark}

\author{%
  Le Yu\\
  Beihang University\\
  \texttt{yule@buaa.edu.cn}\\
}

\begin{document}

\maketitle

\begin{abstract}
In this paper, we conduct an empirical evaluation of Temporal Graph Benchmark (TGB) \cite{huang2023temporal} by extending our Dynamic Graph Library (DyGLib) \cite{yu2023towards} to TGB. Compared with \cite{huang2023temporal}, we include eleven popular dynamic graph learning methods for more exhaustive comparisons. Through the experiments, we find that (\romannumeral1) different models depict varying performance across various datasets, which is in line with the observations in \cite{yu2023towards,huang2023temporal}; (\romannumeral2) the performance of some baselines can be significantly improved over the reported results in \cite{huang2023temporal} when using DyGLib \cite{yu2023towards}. This work aims to ease the researchers' efforts in evaluating various dynamic graph learning methods on TGB and attempts to offer results that can be directly referenced in the follow-up research. All the used resources in this project are publicly available 
at \url{https://github.com/yule-BUAA/DyGLib\_TGB}. This work is in progress, and feedback from the community is welcomed for improvements.
\end{abstract}

\input{section-1-preliminaries}

\input{section-2-experiments}

\input{section-3-conclusion}

\bibliographystyle{plain}
\bibliography{reference.bib}
%%%%%%%%%%%%%%%%%%%%%%%%%%%%%%%%%%%%%%%%%%%%%%%%%%%%%%%%%%%%

% \clearpage
\input{Appendix}

\end{document}

%% file: section-1-preliminaries.tex
\section{Preliminaries}
\label{section-1}

\textbf{Background}. Dynamic graph learning has attracted the increasing attention of many scholars \cite{DBLP:journals/jmlr/KazemiGJKSFP20}. To foster the development of dynamic graph learning, Dynamic Graph Library (DyGLib) \cite{yu2023towards} and Temporal Graph Benchmark (TGB) \cite{huang2023temporal} were introduced very recently. Concretely, DyGLib is a dynamic graph learning library that integrates several popular dynamic graph learning methods in a standard training pipeline. TGB presents a collection of challenging and diverse benchmark datasets for temporal graphs\footnote{In this paper, "dynamic" and "temporal" are used interchangeably.}.

\textbf{Our work}. This work extends DyGLib to TGB to provide more comprehensive and reliable experimental results for the follow-up research. We evaluate eleven methods on dynamic link property prediction and dynamic node property prediction tasks in TGB through a well-coded implementation. We also discuss the key observations in the experiments.

%% file: section-2-experiments.tex
\section{Experiments}
\label{section-2}

\subsection{Experimental Settings}
We closely follow the experimental settings in DyGLib \cite{yu2023towards} and TGB \cite{huang2023temporal}. Our project is developed based on the repositories of DyGLib\footnote{\url{https://github.com/yule-BUAA/DyGLib}} and TGB\footnote{\url{https://github.com/shenyangHuang/TGB}}.

\textbf{Datasets, Baselines, and Evaluation Metrics}. We conduct the experiments on TGB, including five datasets (i.e., tgbl-wiki, tgbl-review, tgbl-coin, tgbl-comment, and tgbl-flight) for the dynamic link property prediction task, as well as three datasets (i.e., tgbn-trade, tgbn-genre, and tgbn-reddit) for the dynamic node property prediction task. We report the performance of nine methods (i.e., JODIE \cite{DBLP:conf/kdd/KumarZL19}, DyRep \cite{DBLP:conf/iclr/TrivediFBZ19}, TGAT \cite{DBLP:conf/iclr/XuRKKA20}, TGN \cite{DBLP:journals/corr/abs-2006-10637}, CAWN \cite{DBLP:conf/iclr/WangCLL021}, EdgeBank \cite{poursafaei2022towards}, TCL \cite{DBLP:journals/corr/abs-2105-07944}, GraphMixer \cite{cong2023do}, and DyGFormer \cite{yu2023towards}) on the dynamic link property prediction task, and evaluate ten methods (i.e., JODIE \cite{DBLP:conf/kdd/KumarZL19}, DyRep \cite{DBLP:conf/iclr/TrivediFBZ19}, TGAT \cite{DBLP:conf/iclr/XuRKKA20}, TGN \cite{DBLP:journals/corr/abs-2006-10637}, CAWN \cite{DBLP:conf/iclr/WangCLL021}, TCL \cite{DBLP:journals/corr/abs-2105-07944}, GraphMixer \cite{cong2023do}, DyGFormer \cite{yu2023towards}, Persistent Forecast \cite{salcedo2022persistence}, and Moving Average \cite{panch2018artificial}) on the dynamic node property prediction task. Mean Reciprocal Rank (MRR) is used to evaluate dynamic link property prediction and Normalized Discounted Cumulative Gain (NDCG) is computed for measuring dynamic node property prediction. We direct interested readers to the above references for further details of the datasets and methods.

\textbf{Model Configurations}. We perform the grid search on the small-scale tgbl-wiki and tgbn-trade datasets to find the optimal configurations of several critical hyperparameters of various methods. Due to the limitations of computational resources, we treat the optimal hyperparameters of baselines on tgbl-wiki/tgbn-trade as the default setting and apply it to other dynamic link/node property prediction datasets. More comprehensive selections of the hyperparameters are expected in the future. Please see \secref{section-appendix-configurations} for details of the grid search as well as the detailed configurations of various models.

\textbf{Implementation Details}. Following \cite{yu2023towards}, we use Adam as the optimizer. We train models that contain learnable parameters (i.e., excluding EdgeBank, Persistent Forecast, and Moving Average) for 100 epochs and use the early stopping strategy with the patience of 20 or 10 (used when the model training speed is slow). The model that performs best on the validation set is selected for testing. We set the learning rate and batch size to 0.0001 and 200 for all the methods on all the datasets. We run the methods five times and report the average performance to eliminate deviations. The experiments are conducted on an NVIDIA GeForce RTX 3090 with 24 GB memory, and an NVIDIA Tesla V100 GPU with 32 GB memory is used for models with more computational costs.

\subsection{Performance for Dynamic Link Property Prediction}
We report the performance of different methods for dynamic link property prediction in \tabref{tab:mrr_dynamic_link_property_prediction}. The best and second-best results are emphasized by \textbf{bold} and \underline{underlined} fonts. The results are multiplied by 100 for a better display layout. Note that some results are missing since the models either took too long to train or raised an out-of-memory issue on GPU. Numbers in \textcolor{red}{red} mean that we cannot obtain results due to computational cost issues and instead use the numbers reported in \cite{huang2023temporal} (although some of them may be not so rigorous).

\begin{table}[!htbp]
\centering
\caption{MRR for dynamic link property prediction, where Val is the abbreviation of Validation.}
\label{tab:mrr_dynamic_link_property_prediction}
\resizebox{1.01\textwidth}{!}
{
\setlength{\tabcolsep}{0.5mm}
{
\begin{tabular}{c|c|ccccc}
\hline
        Sets                     & Methods                 & tgbl-wiki-v2  & tgbl-review-v2  & tgbl-coin-v2 & tgbl-comment & tgbl-flight-v2 \\ \hline
\multirow{12}{*}{Val} & JODIE                   &  71.42 $\pm$ 0.76          &  \textbf{34.76 $\pm$ 0.06}    &           &              &             \\
                             & DyRep                   &  59.38 $\pm$ 1.82          &  \underline{33.85 $\pm$ 0.18}  &    \textcolor{red}{51.20 $\pm$ 1.40}         &    \textcolor{red}{29.10 $\pm$ 2.80}          &    \underline{\textcolor{red}{57.30 $\pm$ 1.30}}         \\
                             & TGAT                   &  65.14 $\pm$ 1.22    &    17.24 $\pm$ 0.89       &   60.47 $\pm$ 0.22        &     50.73 $\pm$ 2.47         &             \\
                             & TGN                    &  73.80 $\pm$ 0.39        &     33.17 $\pm$ 0.13    &  \textcolor{red}{60.70 $\pm$ 1.40}         &    \textcolor{red}{35.60 $\pm$ 1.90}          &     \textbf{\textcolor{red}{73.10 $\pm$ 1.00}}        \\
                             & CAWN                   &  75.36 $\pm$ 0.34       &     \textcolor{red}{20.00 $\pm$ 0.10}      &           &              &             \\
                             & EdgeBank$_\infty$       &  56.13 $\pm$ 0.00  &  2.29 $\pm$ 0.00  &      31.54 $\pm$ 0.00       &    \textcolor{red}{10.87 $\pm$ 0.00}          &    \textcolor{red}{16.60 $\pm$ 0.00}         \\
                             & EdgeBank$_\text{tw-ts}$  &  66.51 $\pm$ 0.00  &  2.90 $\pm$ 0.00  &     49.67 $\pm$ 0.00      &     \textcolor{red}{12.44 $\pm$ 0.00}         &    \textcolor{red}{36.30 $\pm$ 0.00}         \\
                             & EdgeBank$_\text{tw-re}$  &  68.82 $\pm$ 0.00  &  2.98 $\pm$ 0.00  &    54.51 $\pm$ 0.00        &              &             \\
                             & EdgeBank$_\text{th}$   &  53.77 $\pm$ 0.00   &  1.98 $\pm$ 0.00  &     41.99 $\pm$ 0.00       &              &             \\
                             & TCL                   &  \underline{80.82 $\pm$ 0.14}   &   17.99 $\pm$ 1.72     &     66.85 ± 0.27        &      \underline{65.10 $\pm$ 0.67}       &             \\
                             & GraphMixer          &  63.87 $\pm$ 0.53        &    28.28 $\pm$ 2.07       &    \underline{70.38 $\pm$ 0.40}        &     \textbf{70.19 $\pm$ 0.23}         &             \\
                             & DyGFormer             &  \textbf{81.62 $\pm$ 0.46}       &    21.92 $\pm$ 1.74      &    \textbf{72.97 $\pm$ 0.23}       &      61.33 $\pm$ 0.27        &             \\ \hline
\multirow{12}{*}{Test}       & JODIE                 &  63.05 $\pm$ 1.69        &   \textbf{41.43 $\pm$ 0.15}     &           &              &             \\
                             & DyRep                  &  51.91 $\pm$ 1.95     &     \underline{40.06 $\pm$ 0.59}      &     \textcolor{red}{45.20 $\pm$ 4.60}      &    \textcolor{red}{28.90 $\pm$ 3.30}          &     \underline{\textcolor{red}{55.60 $\pm$ 1.40}}        \\
                             & TGAT                   &    59.94 $\pm$ 1.63    &    19.64 $\pm$ 0.23       &   60.92 $\pm$ 0.57        &    56.20 $\pm$ 2.11          &             \\
                             & TGN                     &  68.93 $\pm$ 0.53       &    37.48 $\pm$ 0.23       &    \textcolor{red}{58.60 $\pm$ 3.70}        &   \textcolor{red}{37.90 $\pm$ 2.10}            &    \textbf{\textcolor{red}{70.50 $\pm$ 2.00}}         \\
                             & CAWN                     &  73.04 $\pm$ 0.60      &     \textcolor{red}{19.30 $\pm$ 0.10}      &           &              &             \\
                             & EdgeBank$_\infty$      &  52.50 $\pm$ 0.00  &  2.29 $\pm$ 0.00  &      35.90 $\pm$ 0.00       &   \textcolor{red}{12.85 $\pm$ 0.00}           &   \textcolor{red}{16.70 $\pm$ 0.00}          \\
                             & EdgeBank$_\text{tw-ts}$ &  63.25 $\pm$ 0.00   &  2.94 $\pm$ 0.00  &      57.36 $\pm$ 0.00      &   \textcolor{red}{14.94 $\pm$ 0.00}           &   \textcolor{red}{38.70 $\pm$ 0.00}          \\
                             & EdgeBank$_\text{tw-re}$  &  65.88 $\pm$ 0.00    &  2.84 $\pm$ 0.00  &    59.15  $\pm$ 0.00        &              &             \\
                             & EdgeBank$_\text{th}$    &  52.81 $\pm$ 0.00   &  1.97 $\pm$ 0.00  &       43.36 $\pm$ 0.00       &              &             \\
                             & TCL                    &  \underline{78.11 $\pm$ 0.20}  &  16.51 $\pm$ 1.85  &     68.66 $\pm$ 0.30          &    \underline{70.11 $\pm$ 0.83}        &             \\
                             & GraphMixer              &  59.75 $\pm$ 0.39      &     36.89 $\pm$ 1.50     &   \textbf{75.57 $\pm$ 0.27}         &      \textbf{76.17 $\pm$ 0.17}        &             \\
                             & DyGFormer               &  \textbf{79.83 $\pm$ 0.42}     &     22.39 $\pm$ 1.52     &    \underline{75.17 $\pm$ 0.38}       &      67.03 $\pm$ 0.14        &             \\ \hline
\end{tabular}
}
}
\end{table}

From \tabref{tab:mrr_dynamic_link_property_prediction}, we have two key observations. 
Firstly, different methods show varying performance on different datasets, demonstrating their unique characteristics as well as the diversity of datasets. Moreover, trainable parametric methods can always perform better than the non-parametric EdgeBank, which indicates the complexity of the dynamic link property prediction task. Secondly, compared with \cite{huang2023temporal}, the performance of some baselines (e.g., DyRep, TGN, TCL, and GraphMixer) can be effectively promoted when using DyGLib as the backbone, which proves the importance of introducing a unified library for dynamic graph learning methods \cite{yu2023towards}.

\subsection{Performance for Dynamic Node Property Prediction}
We report the results of dynamic node property prediction in \tabref{tab:ndcg_dynamic_node_property_prediction}. We find that (\romannumeral1) simple non-trainable Persistent Forecast and Moving Average methods can achieve surprisingly good performance than other trainable methods; (\romannumeral2) current popular dynamic graph learning methods mainly focus on link-based tasks and cannot obtain satisfactory results on node-based tasks, which indicates the necessity of designing specialized models for the dynamic node property prediction task.

\begin{table}[!htbp]
\caption{NDCG for dynamic node property prediction.}
\label{tab:ndcg_dynamic_node_property_prediction}
\resizebox{1.01\textwidth}{!}
{
\setlength{\tabcolsep}{1.0mm}
{
\begin{tabular}{c|cc|cc|cc}
\hline
Methods             & \multicolumn{2}{c|}{tgbn-trade} & \multicolumn{2}{c|}{tgbn-genre} & \multicolumn{2}{c}{tgbn-reddit} \\ \hline
                    & Validation     & Test           & Validation     & Test           & Validation        & Test        \\ \hline
JODIE               & 39.35 $\pm$ 0.05   & 37.43 $\pm$ 0.09   &    35.79 $\pm$ 0.03            &    34.99 $\pm$ 0.04            &     34.64 $\pm$ 0.02              &    31.39 $\pm$ 0.01         \\
DyRep               & 39.48 $\pm$ 0.14   & 37.50 $\pm$ 0.07   &     36.00 $\pm$ 0.04           &    35.22 $\pm$ 0.03            &        34.64 $\pm$ 0.01           &    31.40 $\pm$ 0.01        \\
TGAT                & 39.31 $\pm$ 0.01   & 37.40 $\pm$ 0.06   &      35.77 $\pm$ 0.01          &    34.95 $\pm$ 0.01            &      34.65 $\pm$ 0.01             &     31.40 $\pm$ 0.01        \\
TGN                 & 39.35 $\pm$ 0.07   & 37.39 $\pm$ 0.06   &      35.92 $\pm$ 0.06          &    35.18 $\pm$ 0.05            &      34.67 $\pm$ 0.02          &      	31.42 $\pm$ 0.01       \\
CAWN                &    39.28 $\pm$ 0.07            &  37.35 $\pm$ 0.09              &                &                &                   &             \\
TCL                 & 39.41 $\pm$ 0.11   & 37.46 $\pm$ 0.09   &     36.23 $\pm$ 0.04           &   35.44 $\pm$ 0.02             &  34.68 $\pm$ 0.01                 &     31.43 $\pm$ 0.01         \\
GraphMixer          & 39.44 $\pm$ 0.17   & 37.47 $\pm$ 0.11   &     36.06 $\pm$ 0.04          &     35.23 $\pm$ 0.03           &    34.67 $\pm$ 0.01               &    31.42 $\pm$ 0.01         \\
DyGFormer           & 40.77 $\pm$ 0.58   & 38.78 $\pm$ 0.64   &     \underline{37.09 $\pm$ 0.06}           &    \underline{36.51 $\pm$ 0.20}            &      34.84 $\pm$ 0.02             &    31.56 $\pm$ 0.01         \\
Persistent Forecast & \textbf{86.72 $\pm$ 0.00}   & \textbf{84.69 $\pm$ 0.00}   & 35.24 $\pm$ 0.00   & 35.96 $\pm$ 0.00   &     \underline{37.99 $\pm$ 0.00}           & \underline{37.72 $\pm$ 0.00}          \\
Moving Average      & \underline{84.25 $\pm$ 0.00}   & \underline{80.93 $\pm$ 0.00}   & \textbf{50.74 $\pm$ 0.00}   & \textbf{51.79 $\pm$ 0.00}   &    \textbf{57.41 $\pm$ 0.00}             &    \textbf{56.24 $\pm$ 0.00}     \\ \hline
\end{tabular}
}
}
\end{table}

%% file: section-3-conclusion.tex
\section{Conclusion}
\label{section-3}
In this work, we provided an extension of DyGLib to TGB and offered an empirical evaluation of TGB. By conducting the experiments, we presented comprehensive results of several existing dynamic graph learning methods. Our goal is to reduce the efforts of researchers in evaluating different dynamic graph learning methods and facilitates the development of new approaches. This work is ongoing and we welcome feedback from the community to improve it further.

%% file: Appendix.tex
\appendix
\label{section-appendix}

\section{Configurations of Various Methods}\label{section-appendix-configurations}
\tabref{tab:searched_ranges_related_methods} shows the searched ranges and related methods in the grid search.
 
\begin{table}[!htbp]
\centering
\caption{Searched ranges of hyperparameters and the related methods.}
\label{tab:searched_ranges_related_methods}
\setlength{\tabcolsep}{2.0mm}
{
\begin{tabular}{c|cc}
\hline
Hyperparameters                                                             & Searched Ranges                                                                                                                        & Related Methods                                                                                      \\ \hline
Dropout Rate                                                              & \begin{tabular}[c]{@{}c@{}}[0.0, 0.1, 0.2, 0.3, \\ 0.4, 0.5, 0.6]\end{tabular}                                                         & \begin{tabular}[c]{@{}c@{}}JODIE, DyRep, TGAT, TGN, CAWN, \\ TCL, GraphMixer, DyGFormer\end{tabular} \\
\begin{tabular}[c]{@{}c@{}}Number of \\ Sampled Neighbors\end{tabular}      & [10, 20, 30]                                                                                                                           & \begin{tabular}[c]{@{}c@{}}DyRep, TGAT, TGN, \\ TCL, GraphMixer\end{tabular}                         \\
\begin{tabular}[c]{@{}c@{}}Neighbor Sampling \\ Strategies\end{tabular}     & [uniform,recent]                                                                                                                       & \begin{tabular}[c]{@{}c@{}}DyRep, TGAT, TGN, \\ TCL, GraphMixer\end{tabular}                         \\
\begin{tabular}[c]{@{}c@{}}Number of Causal \\ Anonymous Walks\end{tabular} & [32, 64]                                                                                                                      & CAWN                                                                                                 \\
\begin{tabular}[c]{@{}c@{}}Length of Input \\ Sequences \& \\ Patch Size\end{tabular} & \begin{tabular}[c]{@{}c@{}}[32 \& 1, 64 \& 2, 128 \& 4, \\ 256 \& 8, 512 \& 16]\end{tabular}   & DyGFormer                                                                                            \\ \hline
\end{tabular}
}
\end{table}

We show the configurations of different methods as follows, where DLPP and DNPP are abbreviations for Dynamic Link Property Prediction and Dynamic Node Property Prediction. Note that when the settings of DLPP and DNPP are different, we list them separately, otherwise, we only list one item.

\begin{itemize}
    \item  \textbf{JODIE}:
    \begin{itemize}
    \item Dimension of time encoding: 100
    \item Dimension of node memory: 172
    \item Dimension of output representation: 172
    \item Memory updater: vanilla recurrent neural network
    \item Dropout rate: 0.1
    \end{itemize}

    \item  \textbf{DyRep}:
    \begin{itemize}
    \item Dimension of time encoding: 100
    \item Dimension of node memory: 172
    \item Dimension of output representation: 172
    \item Number of graph attention heads: 2
    \item Number of graph convolution layers: 1
    \item Memory updater: vanilla recurrent neural network
    \item Dropout rate: 0.1 on DLPP, 0.0 on DNPP
    \item Number of sampled neighbors: 10
    \item Neighbor sampling strategy: recent
    \end{itemize}
    
    \item  \textbf{TGAT}:
    \begin{itemize}
    \item Dimension of time encoding: 100
    \item Dimension of output representation: 172
    \item Number of graph attention heads: 2
    \item Number of graph convolution layers: 2
    \item Dropout rate: 0.1 on DLPP, 0.2 on DNPP
    \item Number of sampled neighbors: 20
    \item Neighbor sampling strategy: recent
    \end{itemize}
    
    \item  \textbf{TGN}:
    \begin{itemize}
    \item Dimension of time encoding: 100
    \item Dimension of node memory: 172
    \item Dimension of output representation: 172
    \item Number of graph attention heads: 2
    \item Number of graph convolution layers: 1
    \item Memory updater: gated recurrent unit
    \item Dropout rate: 0.1 on DLPP, 0.0 on DNPP
    \item Number of sampled neighbors: 10
    \item Neighbor sampling strategy: recent
    \end{itemize}
    
    \item  \textbf{CAWN}:
    \begin{itemize}
    \item Dimension of time encoding: 100
    \item Dimension of position encoding: 172
    \item Dimension of output representation: 172
    \item Number of attention heads for encoding walks: 8
    \item Length of each walk (including the target node): 2
    \item Time scaling factor $\alpha$: 1e-6
    \item Dropout rate: 0.1
    \item Number of causal anonymous walks: 32
    \end{itemize}
    
    \item  \textbf{TCL}:
    \begin{itemize}
    \item Dimension of time encoding: 100
    \item Dimension of depth encoding: 172
    \item Dimension of output representation: 172
    \item Number of attention heads: 2
    \item Number of Transformer layers: 2
    \item Dropout rate: 0.1 on DLPP, 0.0 on DNPP
    \item Number of sampled neighbors: 20
    \item Neighbor sampling strategy: recent
    \end{itemize}

    \item  \textbf{GraphMixer}:
    \begin{itemize}
    \item Dimension of time encoding: 100
    \item Dimension of output representation: 172
    \item Number of MLP-Mixer layers: 2
    \item Time gap $T$: 2000
    \item Dropout rate: 0.5 on DLPP, 0.1 on DNPP
    \item Number of sampled neighbors: 30
    \item Neighbor sampling strategy: recent
    \end{itemize}

    \item  \textbf{DyGFormer}:
    \begin{itemize}
    \item Dimension of time encoding: 100
    \item Dimension of neighbor co-occurrence encoding: 50
    \item Dimension of aligned encoding: 50
    \item Dimension of output representation: 172
    \item Number of attention heads: 2
    \item Number of Transformer layers: 2
    \item Dropout rate: 0.1
    \item Length of input sequences: 32 on DLPP, 256 on DNPP
    \item Patch size: 1 on DLPP, 8 on DNPP
    \end{itemize}

    \item  \textbf{Moving Average}:
    \begin{itemize}
    \item Window size: 7
    \end{itemize}  
\end{itemize}

%% file: main.bbl
\begin{thebibliography}{10}

\bibitem{cong2023do}
Weilin Cong, Si~Zhang, Jian Kang, Baichuan Yuan, Hao Wu, Xin Zhou, Hanghang
  Tong, and Mehrdad Mahdavi.
\newblock Do we really need complicated model architectures for temporal
  networks?
\newblock In {\em International Conference on Learning Representations}, 2023.

\bibitem{huang2023temporal}
Shenyang Huang, Farimah Poursafaei, Jacob Danovitch, Matthias Fey, Weihua Hu,
  Emanuele Rossi, Jure Leskovec, Michael Bronstein, Guillaume Rabusseau, and
  Reihaneh Rabbany.
\newblock Temporal graph benchmark for machine learning on temporal graphs.
\newblock {\em arXiv preprint arXiv:2307.01026}, 2023.

\bibitem{DBLP:journals/jmlr/KazemiGJKSFP20}
Seyed~Mehran Kazemi, Rishab Goel, Kshitij Jain, Ivan Kobyzev, Akshay Sethi,
  Peter Forsyth, and Pascal Poupart.
\newblock Representation learning for dynamic graphs: {A} survey.
\newblock {\em J. Mach. Learn. Res.}, 21:70:1--70:73, 2020.

\bibitem{DBLP:conf/kdd/KumarZL19}
Srijan Kumar, Xikun Zhang, and Jure Leskovec.
\newblock Predicting dynamic embedding trajectory in temporal interaction
  networks.
\newblock In {\em Proceedings of the 25th {ACM} {SIGKDD} International
  Conference on Knowledge Discovery {\&} Data Mining, {KDD} 2019}, pages
  1269--1278. {ACM}, 2019.

\bibitem{panch2018artificial}
Trishan Panch, Peter Szolovits, and Rifat Atun.
\newblock Artificial intelligence, machine learning and health systems.
\newblock {\em Journal of global health}, 8(2), 2018.

\bibitem{poursafaei2022towards}
Farimah Poursafaei, Andy Huang, Kellin Pelrine, and Reihaneh Rabbany.
\newblock Towards better evaluation for dynamic link prediction.
\newblock In {\em Thirty-sixth Conference on Neural Information Processing
  Systems Datasets and Benchmarks Track}, 2022.

\bibitem{DBLP:journals/corr/abs-2006-10637}
Emanuele Rossi, Ben Chamberlain, Fabrizio Frasca, Davide Eynard, Federico
  Monti, and Michael~M. Bronstein.
\newblock Temporal graph networks for deep learning on dynamic graphs.
\newblock {\em CoRR}, abs/2006.10637, 2020.

\bibitem{salcedo2022persistence}
S~Salcedo-Sanz, D~Casillas-P{\'e}rez, J~Del~Ser, C~Casanova-Mateo, L~Cuadra,
  M~Piles, and G~Camps-Valls.
\newblock Persistence in complex systems.
\newblock {\em Physics Reports}, 957:1--73, 2022.

\bibitem{DBLP:conf/iclr/TrivediFBZ19}
Rakshit Trivedi, Mehrdad Farajtabar, Prasenjeet Biswal, and Hongyuan Zha.
\newblock Dyrep: Learning representations over dynamic graphs.
\newblock In {\em 7th International Conference on Learning Representations,
  {ICLR} 2019}. OpenReview.net, 2019.

\bibitem{DBLP:journals/corr/abs-2105-07944}
Lu~Wang, Xiaofu Chang, Shuang Li, Yunfei Chu, Hui Li, Wei Zhang, Xiaofeng He,
  Le~Song, Jingren Zhou, and Hongxia Yang.
\newblock {TCL:} transformer-based dynamic graph modelling via contrastive
  learning.
\newblock {\em CoRR}, abs/2105.07944, 2021.

\bibitem{DBLP:conf/iclr/WangCLL021}
Yanbang Wang, Yen{-}Yu Chang, Yunyu Liu, Jure Leskovec, and Pan Li.
\newblock Inductive representation learning in temporal networks via causal
  anonymous walks.
\newblock In {\em 9th International Conference on Learning Representations,
  {ICLR} 2021}. OpenReview.net, 2021.

\bibitem{DBLP:conf/iclr/XuRKKA20}
Da~Xu, Chuanwei Ruan, Evren K{\"{o}}rpeoglu, Sushant Kumar, and Kannan Achan.
\newblock Inductive representation learning on temporal graphs.
\newblock In {\em 8th International Conference on Learning Representations,
  {ICLR} 2020}. OpenReview.net, 2020.

\bibitem{yu2023towards}
Le~Yu, Leilei Sun, Bowen Du, and Weifeng Lv.
\newblock Towards better dynamic graph learning: New architecture and unified
  library.
\newblock {\em Advances in Neural Information Processing Systems}, 2023.

\end{thebibliography}
